# AGDC: Automatic Garbage Detection and Collection

Siddhant Bansal[1], Seema Patel[1], Ishita Shah[1], Prof. Alpesh Patel[1], Prof. Jagruti Makwana[1], Dr. Rajesh Thakker[1]

[1] Vishwakarma Government Engineering College, Ahmedabad 382424 Gujarat, India

**Abstract.** Waste management is one of the significant problems throughout the world. Contemporaneous methods find it difficult to manage the volume of solid waste generated by the growing urban population. In this paper, we propose a system which is very hygienic and cheap that uses Artificial Intelligence algorithms for detection of the garbage. Once the garbage is detected the system calculates the position of the garbage by the use of the camera only. The proposed system is capable of distinguishing between valuables and garbage with more than 95% confidence in real time. Finally, a robotic arm controlled by the microcontroller is used to pick up the garbage and places it in the bin. Concluding, the paper explains a system that is capable of working as a human in terms of inspecting and collecting the garbage. The system is able to achieve 3-4 frames per second on the Raspberry Pi, capable of detecting the garbage in real time with 90%+ confidence.

## 1. Introduction

India faces major challenges associated with a waste collection[1]. Current Indian system cannot cope up with the volumes of waste generated by the growing urban population and impacts on the environment and public health. Tonnes of garbage are generated all over the world every year[2]. India alone generates approximate 1,00,000 metric tonnes of solid waste every day[2]. Streets and clogged sewers littered with garbage are breeding grounds for diseases and pollution. Government puts the best of their efforts and does house-to-house garbage collection, and in some countries, there is also automation available for the door-to-door collection of the dustbin.[3] But, due to the enormous amount of area, in addition to a house-to-house collection, it is a need of today to have some automation in the field of collection of the garbage lying on the road and open grounds[4]. Talking about the waste collection efficiency, it is the quantity of solid waste collected and deported from streets to disposal sites divided by the total quantity of solid waste generated during the same period[4]. Until now waste collection efficiency is a function of manpower availability, which is hazardous to the people's life also. With the help of automation, this dependency on the manpower can be surmounted as well as it will be hygienic too.

The world is moving towards automation. In the absence of automation and modernization of waste management services, the collection of garbage continues to be labour-intensive activities[5]. Thus, AGDC came up with an automatic garbage collecting system along with adding some intelligence to ordinary garbage collecting robot. Best way to add intelligence to a machine is to use the concept of artificial intelligence. Being more specific, machine learning, which is the subfield of artificial intelligence, focuses on making machines which can learn without being explicitly programmed. AGDC can prove to be a good solution for solid waste management in cities, open grounds, parks, hospitals, shopping malls, and for indoor waste collection as well.

## 2. Overview

AGDC is the robotic system, in which machine learning algorithms are used for the detection of the garbage lying on the ground. After detection of the garbage, the position of the garbage is calculated. This position is shared serially with a microcontroller controlling the robotic arm. A robotic arm collects the detected garbage and puts it into a container which is attached to a robot. Hence, this system can be used as a substitute for humans for finding and collecting the garbage.

The progression of this paper is as follows, in the 1st portion of the methodology, Machine Learning approach for the detection of the garbage is explained. Which includes the description of the object detection and CNN models for the computer vision. In the 2nd portion of the methodology, algorithms for finding the position of the garbage with respect to the camera are described. This location is sent to the microcontroller using serial communication which is explained in the section-3. Now as explained in the section-4 of the methodology, the microcontroller uses that location to control the robotic arm which is used to pick the garbage and place it in the dustbin.

## 3. Methodology

### 3.1 Machine Learning

There are multiple traditional ways in which AGDC can get the garbage detected in the image, but all these methods are not robust and don't have accuracy anywhere near to the humans. To get an accuracy that is what the humans have achieved, AGDC needs to use Machine Learning approach. These approaches have already reached to near human accuracy, and some have also surpassed humans.

**Convolutional Neural Networks**

For solving the problem of object detection, AGDC takes the help of Convolutional Neural Networks (CNN)[6] as they are proven to be very powerful in the field of computer vision. A CNN is a class of deep neural network widely used in image classification, object detection, and all other computer vision tasks. CNN is inspired by the connectivity pattern of the neurons in human and animal visual cortex. The reason why CNN is preferred is that they do not require much rendering of the images whereas; traditional image recognition requires lots of rendering before getting the result. The main difference between traditional algorithms and CNN is that traditional algorithms require hand engineered filters, whereas the CNN learns these features by seeing the images of the same object in a different situation for 'N' number of times. It makes CNN a preferred algorithm as compared to other algorithms.

**Object Detection**

For detecting the garbage from the image, AGDC uses CNN for obtaining the bounding box around the image portion of garbage under test. This is where AGDC performs the task of object detection. Object detection refers to identifying instances of objects of a particular class (such as bottles, cat, dog or truck) in images and videos in digital format. AGDC uses object detection for classifying the garbage with the rest of the objects in the image/video (Fig. 1).

The object detection algorithm enables AGDC to identify places in the image or video where the object of interest i.e. garbage is resting. To make an object detection algorithm for AGDC, pre-trained Mobile Net[7] model is used and letting AGDC in detecting the objects lying in front of the robot. The model was trained to give output in the form of four coordinates of the bounding box around the garbage. This trained model is then transferred to the Raspberry Pi microprocessor board for carrying out the detection process for the robot.

AGDC has few limitations due to pre-trained model. It cannot detect objects for which the MobileNet is not trained. Therefore, in the first prototype, AGDC only detects and considers bottle as a waste material. This makes the model a bit less complicated and allows it to do faster testing. For the future versions, the team is working to train a custom object detection model which will be able to detect all the types of garbage.

The working of the current prototype is that a test image or a frame is sent from the real-time video as an input to the pre-trained CNN and then receive the output in the form of coordinates of the pixels in the image. These coordinates can be used to draw the bounding box for the detected object and visualize the object that is getting detected (Fig. 1).

**Distance and robot movement estimation**

After completing the task of object detection, the next task is to identify the distance of the object from the base of the robotic arm, which is necessary for allowing Robotic arm to pick up the garbage. To complete this task AGDC has found distance with respect to the camera which is used to find the distance with respect to the base of the robotic arm. The task was to design an algorithm that could give the instructions in which direction the robotic arm should move in order to reach the targeted location.

The algorithm developed is divided into two tasks. One for finding the 2D location in front of the camera and second is to find the perpendicular location of the object with respect to the camera.

The first part of the algorithm works as follows. First, the centre of the frame is calculated which is the output of the camera. It's marked as a 'C' in pink colour in Fig. 2. After getting the centre of the frame, the centre of the object is calculated using the coordinates given by CNN. The point is marked as 'C' in blue in Fig. 2. After getting these two points, the PID (Proportional Integral and Derivative control) principle is used to minimize the distance between the 'C' of the object and 'C' of the screen. In the process of minimizing the distance, instructions are given to the robotic arm as 'Move up, right and forward' and when the robot is in the desired location the 'stop' message is sent.

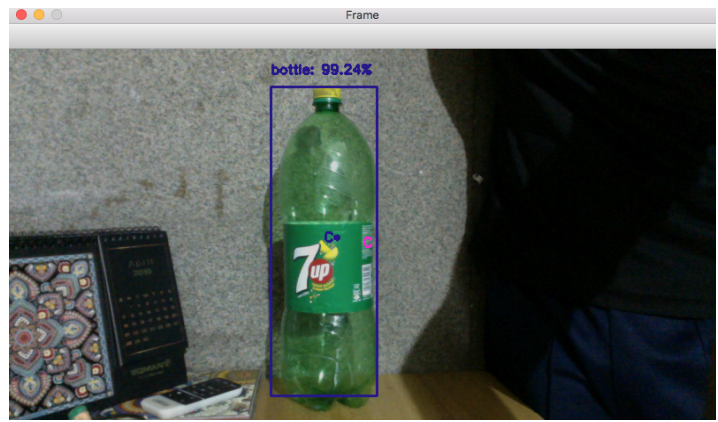

**Fig.1.** Distance approximation demonstration

Once both the centers align, the second task is to find the perpendicular distance of the object from the camera. The following information like the angle of view of the camera, a height of the camera with respect to the ground and tilt angle of the camera. As displayed in the Fig.2. AGDC has a value of the angle other than the right angle in the right-angle triangle and one side. By finding the tan of the angle made by a line joining the camera and the object, it can find the distance d, which AGDC had found the solution for the task.

With this information, the location of the garbage has found with respect to the camera. Now distance from the camera and base of the arm is fixed, AGDC has found the distance of the garbage with respect to the base of the robotic arm.

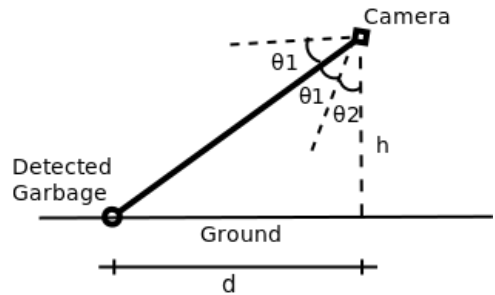

**Fig. 2.** Distance measurement schematic

$$\theta 2 = tilted\ angle\ of\ camera$$
$$h = height\ of\ the\ camera$$
$$d = distance\ of\ the\ garbage\ from\ the\ camera$$
$$2(\theta 1) = view\ angle\ of\ camera$$
$$d = h * \tan(\theta 1 + \theta 2)$$

These inputs are enough to guide the robot in the desired direction and get a hold on the target object. These inputs from the Raspberry Pi are sent to the Arduino via serial communication which in turn moves all the motors accordingly.

### 3.2 Serial Communication

The detection of the object is carried out using machine learning implemented on Raspberry Pi 3b+. Machine learning needs high processing power. A delay is observed if the commands to motors of the robotic arm are given by Raspberry Pi. Thus, the efficiency of picking garbage is decreased. To address this problem, AGDC uses a microcontroller for operating the motors of the arm.

AGDC needs to do serial communication between Raspberry Pi and Arduino Mega controllers. With the help of distance algorithms, the system detects the position of garbage and corresponding commands will be passed to the Arduino controller. The commands from the Raspberry Pi are serially transmitted using the USB protocol. All the commands are obtained on the serial monitor of the Arduino. These commands are read with the help of the Arduino commands and according to the commands the motors are rotated (Fig. 3).

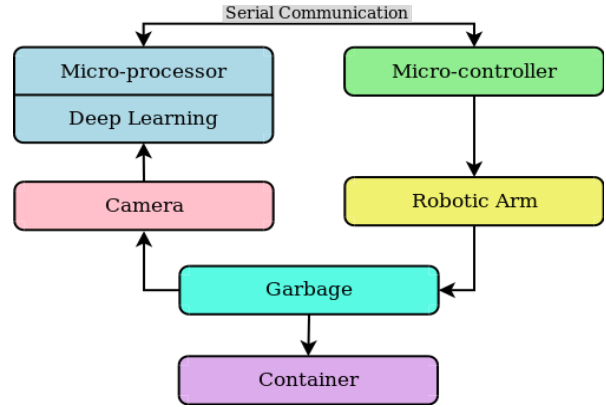

**Fig. 3.** Serial Communication flow

Serial communication is achieved by the means of USB cable. The same can be achieved via I2C or other serial communication protocols but USB cable offers high-speed data transmission as compared to others and ease of connecting two devices as no rebooting is needed for connecting and disconnecting the devices.

### 3.3 Robotic Assembly

After the detection of the garbage, now it is time to pick the garbage in actual. To achieve the same, the robotic arm is developed. The design of the garbage collector can be split into 3 major parts: base, robotic arm, and drawer. The base drives the robot toward the garbage, the robotic arm collects the garbage and the drawer stores the garbage collected by the robotic arm.

**Base**

The base of the AGDC consists of the 4 Johnson geared DC motors with 300 rpm each. These motors are attached to Omni wheels. The reason behind the use of Omni wheels is, with the help of these wheels base can move in all direction (forward, backward, left, right, clockwise, anti-clockwise). Arduino microcontroller board is used to give commands to motors which control the direction to the system. However, these motors need 12V DC volts to be functional, but Arduino can provide a maximum of 5V DC at the output. Hence, motor drivers are used to overcoming this. According to the command received from the Raspberry Pi, Arduino controls the motors and moves the base in given the directions. After reaching the place of garbage, robotic arm comes into picture to pick the garbage and place it in the dustbin.

**Robotic Arm**

The robotic arm is designed in a way to replicate the working of a human arm. The robotic arms are designed to operate wherever human hands aren't able to function. Varied types of Robotic Arms are being designed but three things remain common in each of them: links, joints and, end effectors. The links are individual fixtures which are attached to the joints and are used to form a kinematic chain. The joints are used to give motion to the rigid body(links). (refer fig no 4). The end of this kinematic chain consists of the end effector. The arrangement of the end effector varies according to the application of the arm. It can be any specific designed mechanical part or gripper. In the task of collecting garbage, the gripper is used as an end effector. The algorithm used to reach the garbage is obtained with the help of inverse kinematics. (Refer section on inverse kinematics for more information)

*Mechanical Design*

In the construction of the robotic arm, aluminum plates, aluminum slabs, nuts, bolts, servo motors, and bearings are used. The links of the arm are made up of aluminum plates of sizes 5mm and 3mm. The aluminum is used as it offers high strength and can sustain the load without getting damaged.

The base of the arm is made up of an aluminum slab. The bearing is used in the base to reduce the friction allowing smooth movement of plates while rotation of the entire arm. The angular motion of the arm is achieved using the servo motor fixed in the base.

The servo motors responsible for the shoulder and the elbow movement are connected with the two different vertical aluminum plates which are attached with the main base plate of the arm(refer fig 4). The two links(link0 and link2) are joined to the shaft of the servo motors. The servo motor connected to the arm A (link 2) responsible for its up-down motion is called shoulder servo (refer fig no 4). It behaves like a shoulder joint of the human hand.

The servo motor joined with the small link (link 0) which in turn controls the motion of arm B (link 3) is called the elbow servo. It behaves like an elbow joint of the human hand. If the elbow servo was attached to the shoulder motor via the arm A then it would have to bear the torque of the elbow motor and the links joint to it. So to reduce the torque, the elbow motor is placed on the base plate and the links are joint such that it gives the movement of an elbow to the arm. The small link (link 0) is joined with the one another link (link 1) and this is joined to arm B (refer fig no 4).

The robotic arm is constructed such that the end effector always remains parallel to the ground (refer fig no 4). The four links (link 0, link1, link2, link3) are linked with each other in order to achieve a parallelogram-like structure[8]. This whole structure allows the arm to move in all four directions (up, down, rear and front direction) keeping the gripper parallel to the ground[8].

A servo motor is attached to the end of arm B to provide motion similar to the wrist to the gripper. Another servo motor is used for generating adequate pressure to hold the object tightly. The gripper (Fig.5) is designed using a nut and screw mechanism.

The gripper works using a servo motor, which rotates the bolt and the nut is fixed to the two links of the gripper. As the bolt rotates, the nut is pulled (in and out) allowing the links of the gripper to grab or release an object.

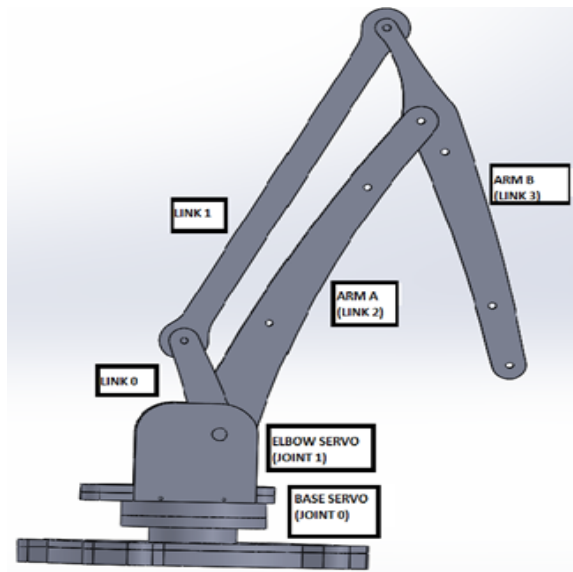

**Fig. 4.** Robotic Arm

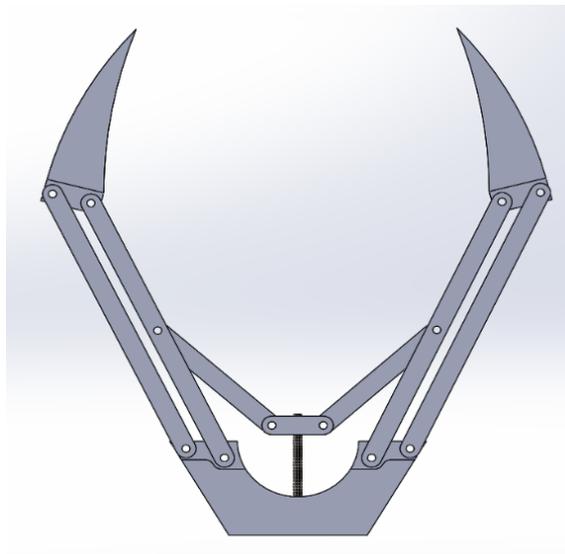

**Fig. 5.** Gripper

*Working of the Arm*

The whole arm is controlled by the 5 servo motors and Arduino board. Servo motors serve as an actuator at various joints. Out of 5 servo motors, the three servo motors are mg955 metal gear and the other two are towerpro sg90 micro servo. All these servos have the feature to rotate 180-degree clockwise direction. As the arm need high stall torque at elbow and shoulder joints to lift the weight of link and object, and the high torque at a base to rotate the whole robotic arm, the servo used must be able to handle the torque of max 7 kg-cm. For that servo motor mg955, which has the ability to handle the 8.5 kg.cm at 4.8V and 10 kg.cm at 6V stall torque are used at the base, elbow and shoulder joints. The wrist joint and motor for gripping the object does not need high torque but should be small in size and less in weight to accomplish this requirement. For that towerpro sg90 servo motor which is 14.7 g and can handle the torque of 2.5 kg.cm at 4.8V are used.

The position of the garbage is detected by the pi and is given to the Arduino board. By obtaining the position in x, y, z coordinates, the Arduino moves each servo motor of the arm to achieve that position in the real world (the centre of the robotic arm is taken as the reference). The movement of a servo to reach to the specific position is done with the help of inverse kinematics (refer section inverse kinematics).

*Degree of Freedom (DOF)*

Usually, the degree of freedom is calculated by the number of actuators used in the arm. The gripper is the most complex part and it needs many actuators. The number of actuators purely depends on the design of gripper thus it is usually avoided while calculating the degree of freedom[9]. This robotic arm of the AGDC has DOF of 4. It can move in all the 3 coordinates (x,y,z) and also the gripper can be rotated just like the wrist of the human hand.

*Robot Workspace*

The robot workspace is a space where the end effector can reach. The workspace depends on the DOF angle/translation limitations, the arm link lengths, the angle at which some object is to be picked up at, etc. The workspace is highly dependent on the robot configuration. With the given design, the robotic arm can approximately move from minimum -46 cm to maximum +46 cm in the x-direction (refer fig no 6), minimum 9 cm to maximum 46 cm in y-direction (refer fig no 6) and minimum 5 cm to maximum 32 cm in z-direction (refer fig no 6), considering the center point of the robotic arm to be origin. Servo motor has less offset value when it is working in the range from 40 degree to 140 degree and have wider offset value when it is working out of the range. Usually, the offset value is around ± 0.5cm.

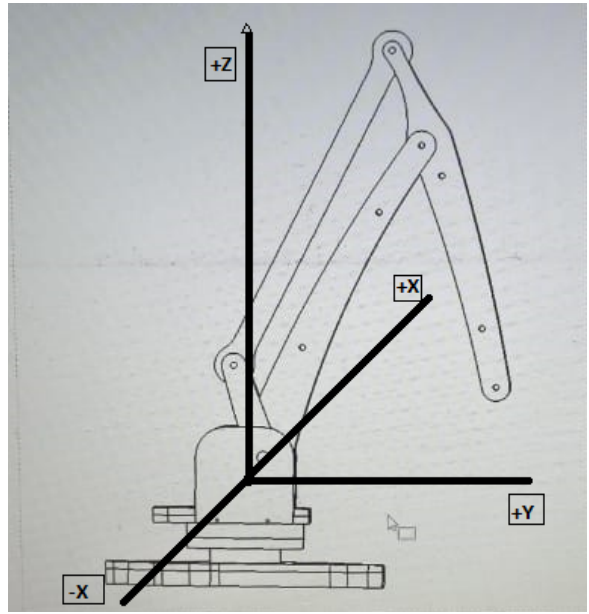

**Fig. 6.** Robotic arm coordinates

*Inverse Kinematic*

The inverse kinematic[10] is the method used to find the angle of the servo motors when the length of the links and the position to achieve is given. As the position of the garbage is obtained by Raspberry Pi, the position to take the gripper is now known and the length of the links (robotic arm) is already defined. Thus by using both these parameters, the angle of the servo motors that are to be displaced could be found[11].

**Power Supply**

The Li-Po battery (3000 mAh) is used to power the Raspberry Pi, Arduino board, servo motors, and Johnson motors. The Li-Po battery supplies 12V maximum. This 12V is provided to the Hercules driver in order to drive the Johnson motor and also to a 12V to 5V converter. The output of this converter is fed to the Raspberry Pi, Arduino and all the servo motors.

## 4. First Prototype of The Robot

The first prototype version made by AGDC can detect any type of bottle garbages with 90%+ confidence. This version can also find the distance of the garbage from the camera with only ±2 cm error, which also can be reduced by properly fixing and finding the tilted angle of the camera.

The robotic arm of this prototype robot was made up of the MDF sheet of dimension 30 x 20 cm and with 3mm thickness. There was three servo motor used and an Arduino board to move the structure as desired. The servo motor used were capable of handling the torque of 2.5 kg.cm. The power to all the servo was given by the Arduino board.

The limitations of this prototype version are that it can lift the garbage with weight up to 100g-200g. The MDF material is not strong enough to take up high loads. If the arm tries to pick objects heavier than the prescribed weight, high torque was generated which would cause the links to break. And another limitation is that it can detect the garbage of bottles only.

## 5. Summary

If we tie all the aforementioned concepts in one system, then it will make automatic garbage detection and collection robot. We just have to set this system in one corner of the ground or at the starting of the road. The camera will start capturing the video and machine learning algorithms will check for the presence of garbage frame by frame of the real-time video. If garbage is detected, then it will find the location of that garbage with respect to itself and communicate the location to the controller. The controller will use inverse kinematics to rotate different servo motors in order to move the gripper at the given location and in turn it will grab the garbage and direct it in the bin attached to the robot itself.

## 6. Future Scope

Current prototype is capable of picking up 100-200gms of garbage. Future version of the robot is being designed keeping in mind to pick up garbage of upto 2-3Kgs in weight. Whereas, the number of garbage that can be detected by the robot can be easily increased either by training a new Convolutional Neural Network from scratch, or by using techniques like transfer learning[12]. The robot can be further connected to the internet and using it as an IOT device.

## 7. Conclusion

This paper introduced a fully automatic system which detects and collects the garbage. We used different concepts for different application to make the whole system. In which pre-trained Convolutional Neural Network[7] is used for detecting the garbage, PID algorithms for detecting the position of the garbage in reference to the camera. We have described the inverse kinematic concept which is used in movement of the robotic arm and nut-and-screw mechanism for a gripper of the robotic arm to pick the garbage. This system is robust and adequately efficient. Overall this robot pushes the state-of-the-art in the waste management in our country with full automation.